\documentclass[11pt,a4paper]{article}
\usepackage[hyperref]{emnlp2020}
\usepackage{times}
\usepackage{latexsym}

\usepackage{microtype}

\usepackage{tabularx,ragged2e}
\usepackage{graphicx}
\usepackage{blindtext}
\usepackage{enumitem}
\usepackage{scalerel}
\usepackage{amsmath}
\usepackage{multicol}
\usepackage{multirow}
\usepackage{booktabs}
\usepackage{arydshln}
\usepackage{float}
\restylefloat{table}

\aclfinalcopy 

\makeatletter
\newcommand{\printfnsymbol}[1]{%
  \textsuperscript{\@fnsymbol{#1}}%
}
\makeatother

\title{MUTANT: A Training Paradigm for Out-of-Distribution Generalization in Visual Question Answering}

\author{
Tejas Gokhale \thanks{~~Equal Contribution} \and
Pratyay Banerjee \printfnsymbol{1}  \and
Chitta Baral \and 
Yezhou Yang
\\ Arizona State University
\\ \texttt{tgokhale,pbanerj6,chitta,yz.yang}@asu.edu
}

\date{}

\begin{document}
\maketitle
\begin{abstract}

While progress has been made on the visual question answering leaderboards, models often utilize spurious correlations and priors in datasets under the i.i.d. setting. 
As such, evaluation on out-of-distribution (OOD) test samples has emerged as a proxy for generalization.
In this paper, we present \textit{MUTANT}, a training paradigm that exposes the model to perceptually similar, yet semantically distinct \textit{mutations} of the input, to improve OOD generalization, such as the VQA-CP challenge.
Under this paradigm, models utilize a consistency-constrained training objective to understand the effect of semantic changes in input (question-image pair) on the output (answer).
Unlike existing methods on VQA-CP, \textit{MUTANT} does not rely on the knowledge about the nature of train and test answer distributions.
\textit{MUTANT} establishes a new state-of-the-art accuracy on VQA-CP with a $10.57\%$ improvement.
Our work opens up avenues for the use of semantic input mutations for OOD generalization in question answering.
\end{abstract}

\section{Introduction}
\begin{figure}[t]
    \centering
    \includegraphics[width=\linewidth]{ 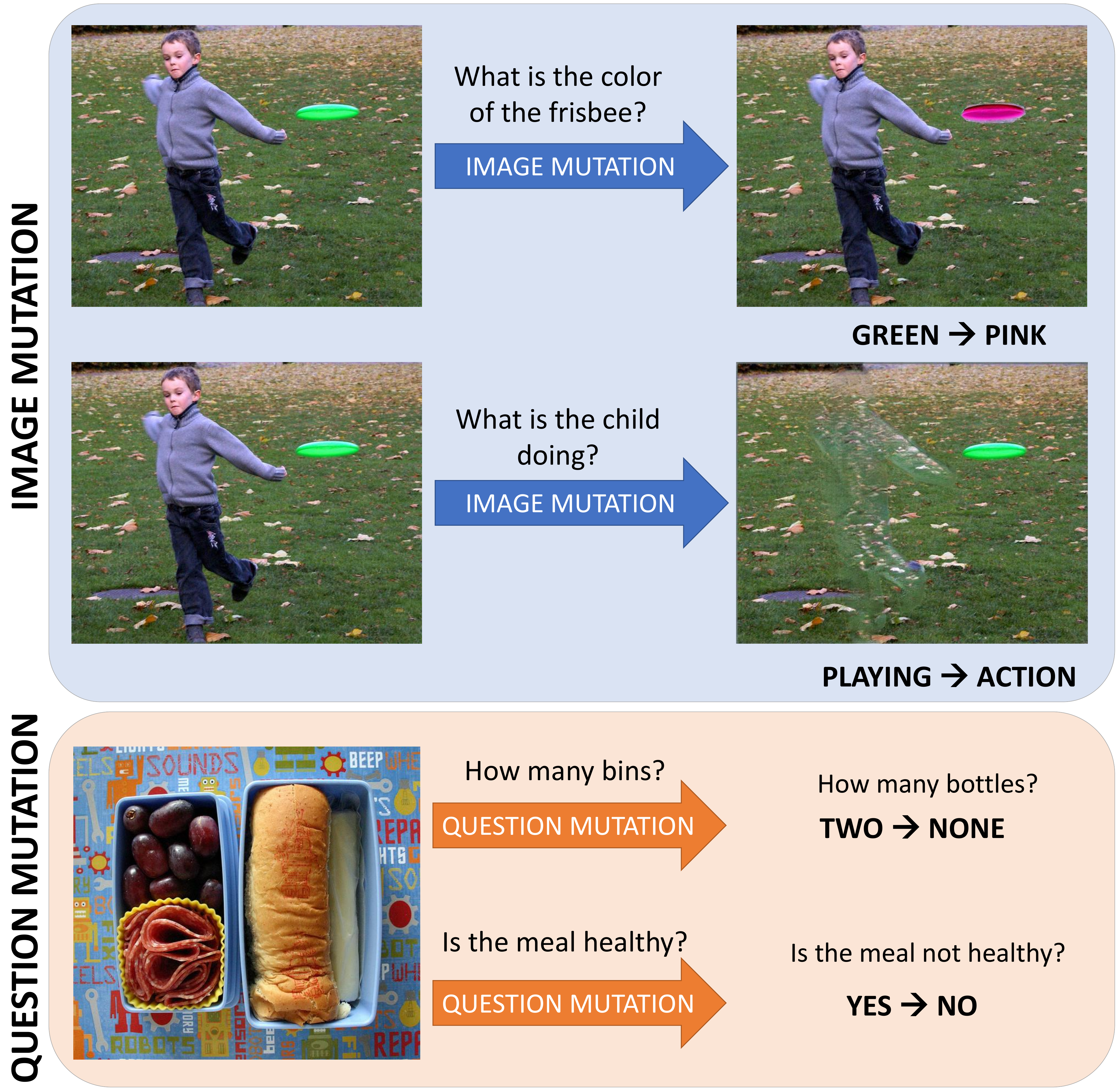}
    \caption{Illustration of the mutant samples. The input mutation, either by manipulating the image or the question, results in a change in the answer.}
    \label{fig:teaser}
\end{figure}

Availability of large-scale datasets has enabled the use of statistical machine learning in vision and language understanding, and has lead to significant advances.
However, the commonly used evaluation criterion is the performance of models on test-samples drawn from the same distribution as the training dataset, which cannot be a measure of generalization.
Training under this ``independent and identically distributed" (i.i.d.)~setting can drive decision making to be highly influenced by dataset biases and spurious correlations as shown in both natural language inference~\cite{kaushik2018much,poliak2018hypothesis,mccoy2019right} and visual question answering~\cite{goyal2017making,vqa-cp,selvaraju2020squinting}.
As such, evaluation on out-of-distribution (OOD) samples has emerged as a metric for generalization.

Visual question answering (VQA)~\cite{antol2015vqa} is a task at the crucial intersection of vision and language.
The aim of VQA models is to provide an answer, given an input image and a question about it.
Large datasets~\cite{antol2015vqa} have been extensively used for developing VQA models.
However over-reliance on datasets can cause models to learn spurious correlations such as linguistic priors~\cite{vqa-cp} that are specific to certain datasets and do not generalize to ``Out-of-Distribution" (OOD) samples, as shown in Figure~\ref{fig:teaser}.
While learning patterns in the data is important, learning dataset-specific spurious correlations is not a feature of robust VQA models.
Developing robust models has thus become a key pursuit for recent work in visual question answering through data augmentation~\cite{goyal2017making}, reorganization~\cite{vqa-cp}.

Every dataset contains biases; indeed inductive bias is \textit{necessary} for machine learning algorithms to work.
\citet{mitchell1980need} states that an unbiased learner's ability to classify is no better than a look-up from memory.
However this bias has a component which is useful for generalization (positive bias), and a component due to spurious correlations (negative bias).
We use the term ``positive bias" to denote the correlations that are necessary to perform a task --- for instance, the answer to a ``What sport is \dots" question is correlated to a name of a sport.
The term ``negative bias" is used for spurious correlations tat may be learned from the data --- for instance, always predicting ``tennis" as the answer to ``What sport\dots" questions.
The goal of OOD generalization is to mitigate negative bias while learning to perform the task.
However existing methods such as LMH~\cite{clark2019don}
try to remove all biases between question-answer pairs, by penalizing examples that can be answered without looking at the image;
we believe this to be counter-productive.
The analogy of antibiotics which are designed to remove pathogen bacteria, but also end up removing useful gut microbiome~\cite{willing2011shifting} is useful to understand this phenomenon.

We present a method that focuses on increasing positive bias and mitigating negative bias, to address the problem of OOD generalization in visual question answering.
Our approach is to enable the \textbf{mutation} of inputs (questions and images) in order to expose the VQA model to perceptually similar yet semantically dissimilar samples. 
The intuition 
is to implicitly allow the model to understand the critical changes in the input which lead to a change in the answer.
This concept of mutations is illustrated in Figure~\ref{fig:teaser}.
If the color of the frisbee is changed, or the child removed, i.e. \textit{when an image-mutation is performed}, the answer to the question changes.
Similarly, if a word is substituted by an adversarial word (bins$\rightarrow$bottles), an antonym, or negation (healthy$\rightarrow$not healthy), i.e. \textit{when a question-mutation is performed}, the answer also changes.
Notice that both mutations do not significantly change the input, most of the pixels in the image and words in the question are unchanged, and the type of reasoning required to answer the question is unchanged.
However the mutation significantly changes the answer.

In this work, we use this concept of mutations to enable models to focus on parts of the input that are critical to the answering process, by training our models to produce answers that are consistent with such mutations.
We present a question-type exposure framework which teaches the model that although such linguistic priors may exist in training data
(such as the dominant answer ``tennis" to ``What sport is ..." questions),
other sports can also be answers to such questions, thus mitigating negative bias.
This is in contrast to \citet{chen2020counterfactual} who focus on using data augmentation as a means for mitigating language bias.

Our method uses a pair-wise training protocol to ensure consistency between answer predictions for the original sample and the mutant sample.
Our model includes a projection layer, which projects cross-modal features and true answers to a learned manifold and uses Noise-Contrastive Estimation Loss~\cite{gutmann2010noise} for minimizing the distance between these two vectors.
Our results establish a new state-of-the-art accuracy of $69.52\%$ on the VQA-CP-v2 benchmark outperforming the current best models by $10.57\%$.
At the same time, our models achieves the best accuracy ($70.24\%$) on VQA-VQA-v2 among models designed for the VQA-CP task.

This work takes a step away from explicit de-biasing as a method for OOD generalization and instead proposes amplification of positive bias and implicit attenuation of spurious correlations as the objective. Our contributions are as follow.
\begin{itemize}[nosep,noitemsep]
    \item We introduce the Mutant paradigm for training VQA models and the sample-generation mechanism which takes advantage of semantic transformations of the input image or question, for the goal of OOD generalization.
    \item In addition to the conventional classification task, we formulate a novel training objective using Noise Contrastive Estimation over the projections of cross-modal features and answer embeddings on a shared projection manifold, to predict the correct answer.
    \item Our pairwise consistency loss acts as a regularization that seeks to bring the distance between ground-truth answer vectors closer to the distance between predicted answer vectors for a pair of original and mutant inputs.
    \item Extensive experiments and analyses demonstrate advantages of our method on the VQA-CP dataset, and establish a new state-of-the-art of $\mathbf{69.52\%}$, an improvement of $\mathbf{10.57\%}$.

\end{itemize}

\section{MUTANT}
We consider the open-ended VQA problem as a multi-class classification problem.
The VQA dataset $\mathcal{D} = \{Q_i, I_i, a_i\}_{i=1}^N$ consists of questions $Q_i \in \mathcal{Q}$ and images $I_i \in {I}$, and answers $a_i \in \mathcal{A}$.
Many contemporary VQA models such as Up-Dn~\cite{anderson2018bottom} and LXMERT~\cite{tan2019lxmert} first extract cross-modal features from the image and question using attention layers, and then use these features as inputs to a neural network answering module which predicts the answer classes.
In this section we define our Mutant paradigm under this formulation of the VQA task.

    \subsection{Concept of Mutations}
    Let $X = (Q, I)$ denote an input to the VQA system with true answer $a$.
    A \textit{mutant} input $X^*$ is created by a small transformation in the image $(Q, I^*)$ or in the question $(Q^*, I)$ such that this transformation leads to a new answer $a^*$, as shown in Figure~\ref{fig:teaser}.
    There are three categories of transformation $T$ that create the mutant input $X^* = T(X)$, addition, removal, or substitution. 
    For image mutations, these correspond to addition or removal of objects, and morphing the attributes of the objects, such as color, texture, and lighting conditions. 
    For instance addition or removal of a person from the image in Figure~\ref{fig:image_mutation} changes the answer to the question ``How many persons are pictured".
    Question mutations can be performed by addition of a negative word (``no", ``not", etc.) to the question, masking critical words in the question, and substituting an object-word with an antonym or adversarial word.
    Thus for each sample in the VQA dataset, we can obtain a mutant sample and use it for training.
    
    \subsection{Training with Mutants}
    Our method of training with mutant samples relies on three key concepts that supplement the conventional VQA classification task.
    
    \paragraph{Answer Projection:}
    
    \begin{figure*}[t]
        \centering
        \includegraphics[width=0.9\linewidth]{ 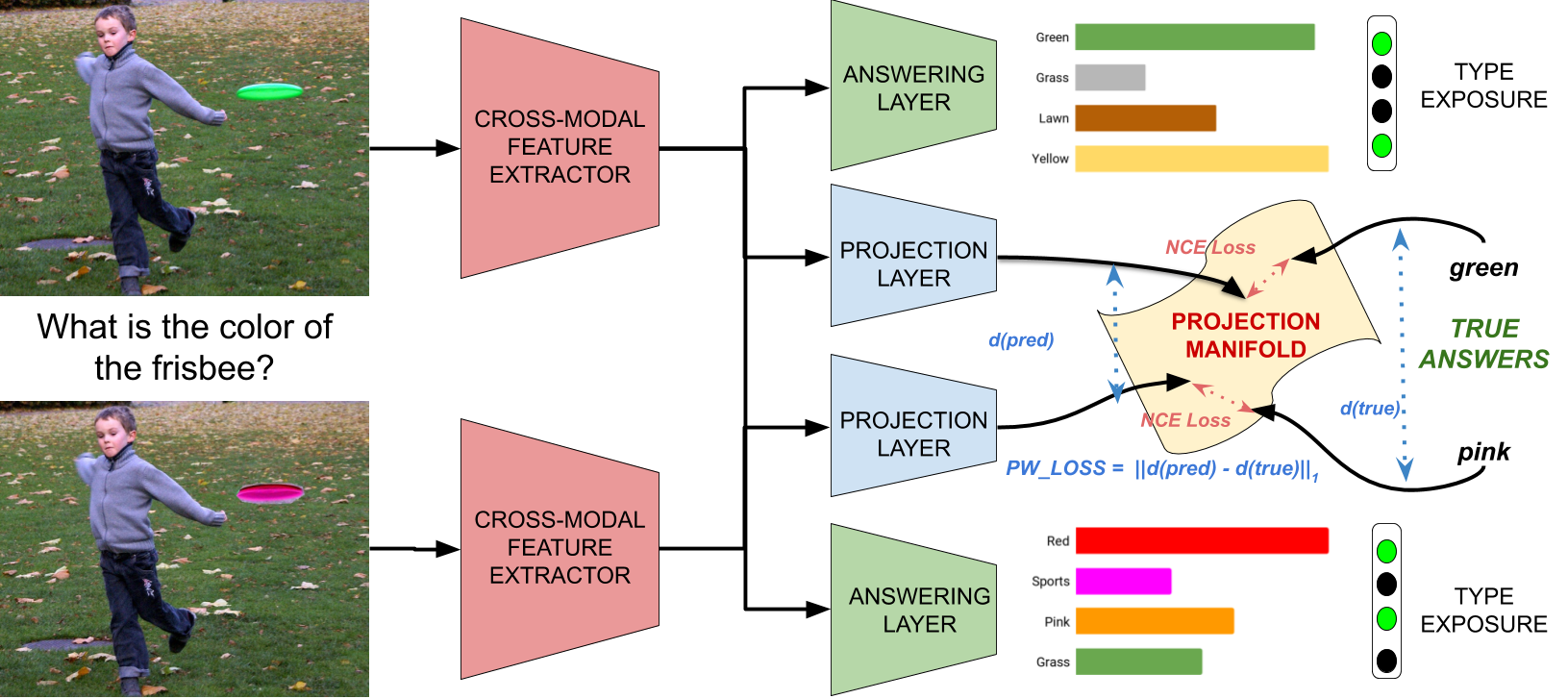}
        \caption{Overall architecture of the Mutant Method includes a cross-modal feature extractor, answer projection layer, answering layer and type exposure model}
        \label{fig:overall}
    \end{figure*}
    
    The traditional learning strategy of VQA models optimizes for a standard classification task using softmax cross-entropy:
    \begin{equation}
        \mathcal{L}_{\scaleto{VQA}{4pt}} = \frac{-1}{N}\sum_{i=1}^{N}\mathit{log}(\mathit{softmax}\big(f_{\mathit{\scaleto{VQA}{4pt}}}(X_i), a_i)).
    \end{equation}
    QA as a classification task is popular since the answer vocabulary follows a long-tailed distribution over the dataset.
    However this formulation is problematic since it does not consider the meaning of the answer while making a decision, but instead learns a correlation between the one-hot vector of the answer-class and input features.
    Thus to answer the question ``What is the color of the banana", models learn a strong correlation between the question features and the answer-class for ``yellow", but do not encode the notion of \textit{yellowness} or \textit{greenness} of bananas.
    This key drawback negatively impacts the generalizability of these models to raw green or over-ripe black bananas at test-time.
    
    To mitigate this, in addition to the classification task, we propose a training objective that operates in the space of answer embeddings.
    The key idea is to map inputs (image-question pairs) and outputs (answers) to a shared manifold in order to establish a metric of similarity on that manifold.
    We train a projection layer that learns to project features and answers to the manifold as shown in Figure~\ref{fig:overall}.
    We then use Noise Contrastive Estimation~\cite{gutmann2010noise} as a loss function to minimize the distance between the projection of cross modal features $z$ and projection of glove vector $v$ for ground-truth answer $\mathit{a}$, given by:
    \begin{equation}
        \mathcal{L}_{\scaleto{NCE}{4pt}} = -log\big(\frac{e^{cos(z_{feat},~~z_{a})}}{\sum_{a_i \in \mathcal{A}} e^{cos(z_{feat},~~z_{a}^i)}}\big),
    \end{equation} 
    where $z_{\scaleto{feat}{5pt}} = f_{\scaleto{proj}{5pt}}(z)$ and $z_{a} = f_{\scaleto{proj}{5pt}}(glove(a))$, and $\mathcal{A}$ is the set of all possible answers in our training dataset.
    It is important to note that this similarity metric is not between the true answer and the predicted answer, but between the projection of input features and the projection of answers, to incorporate context in the answering task.
    
    \paragraph{Type Exposure:}
    Linguistic priors in datasets have led models to learn spurious correlations between question and answers. 
    For instance, in VQA, the most common answer for ``What sport ..." questions is ``tennis", and for ``How many ..." questions is ``two".
    Our aim is to remove this negative bias from the models.
    Instead of removing \textit{all bias} from these models, we teach models to identify the question type, and learn which answers can be valid for a particular question type, irrespective of their frequency of occurrence in the dataset.
    For instance, the answer to ``How many ..." can be all numbers, answers to ``What color ..." can be all colors, and answers to questions such as ``Is the / Are there ..." questions is either yes or no.
    We call this \textit{Type Exposure} since it instructs the model that although a strong correlation may exist between a question-answer pair, there are other answers which are also valid for the specific type of question.
    Our Type Exposure model uses a feedforward network to predict question type and to create a binary mask over answer candidates that correspond to this type.

    \paragraph{Pairwise-Consistency:}
    The final component of Mutant is pairwise consistency.
    We jointly train our models with the original and mutant sample pair, with a loss function that ensures that the distance between two predicted answer vectors is close to the distance between two ground-truth answer vectors.
    The pairwise consistency loss is given below, where $\mathit{z_a}$ is the vector for answer $\mathit{a}$, $\mathit{m}, GT$ denote mutant sample and ground-truth respectively.
    \begin{equation}
        \mathcal{L}_{\scaleto{PW}{4pt}} = ||cos(z_{a_{GT}}, z_{a_{GT}}^m) - cos(z_{a_{pred}}, z_{a_{pred}}^m)||_1 .
        \nonumber
    \end{equation}
    
\begin{figure}[t]
    \centering
    \includegraphics[width=\linewidth]{ 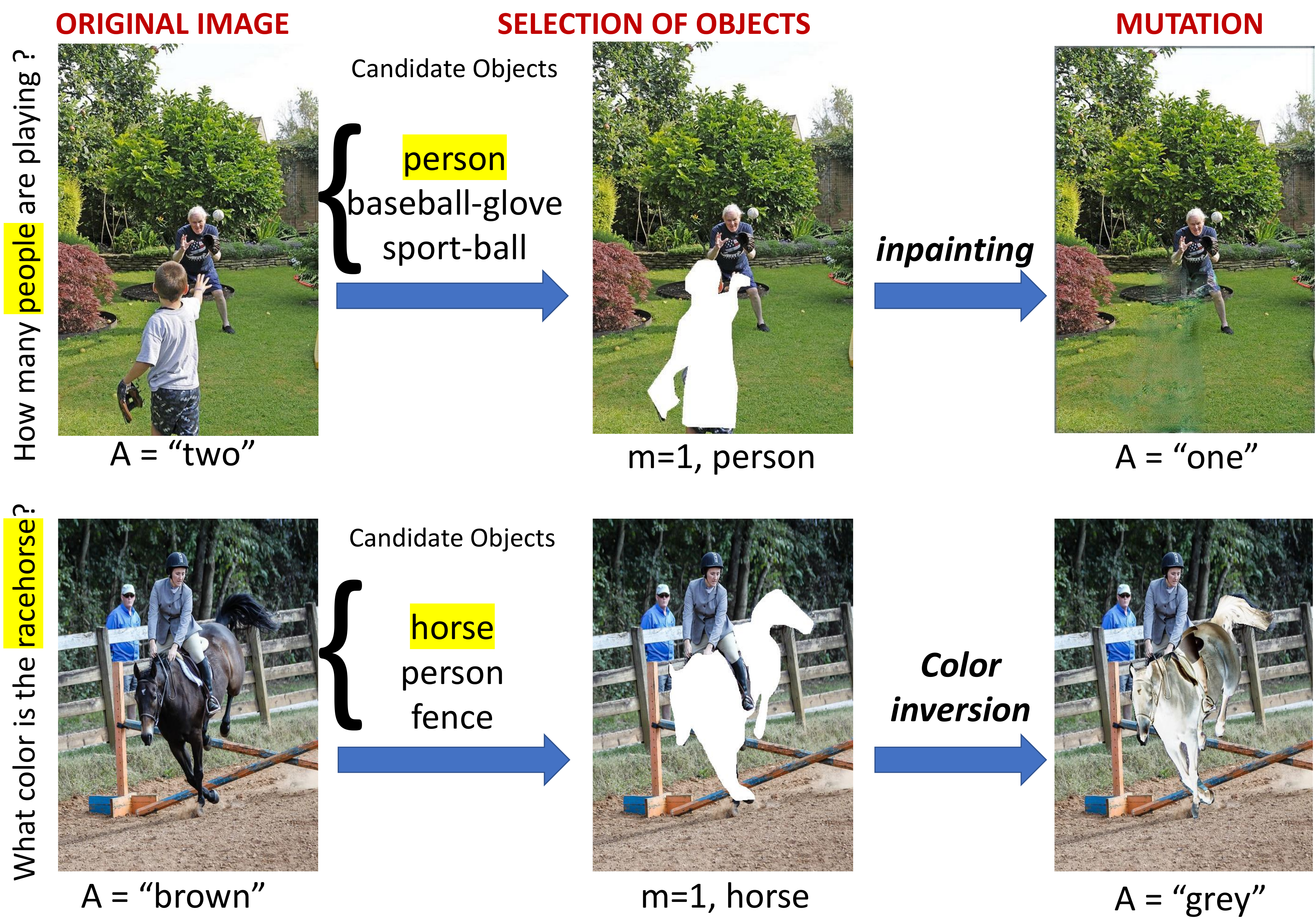}
    \caption{Figure illustrating our dataset creation pipeline for image mutations.
    $m$ object instances of ``critical" object are identified from the question and image, and mutation performed either by removal or color inversion.
    $A$ represents the answer to the question.}
    \label{fig:image_mutation}
\end{figure}

    This pairwise consistency is designed as a regularization that incorporates the notion of semantic shift in answer space as a consequence of a mutation.
    For instance, consider the image mutation in Figure~\ref{fig:image_mutation} which changes the ground-truth answer from "two" to "one".
    This shift in answer-space should be reflected by the predictor.

\section{Generating Input Mutations for VQA}

\begin{figure*}[h]
  \begin{minipage}{.31\textwidth}
    \includegraphics[width=\linewidth]{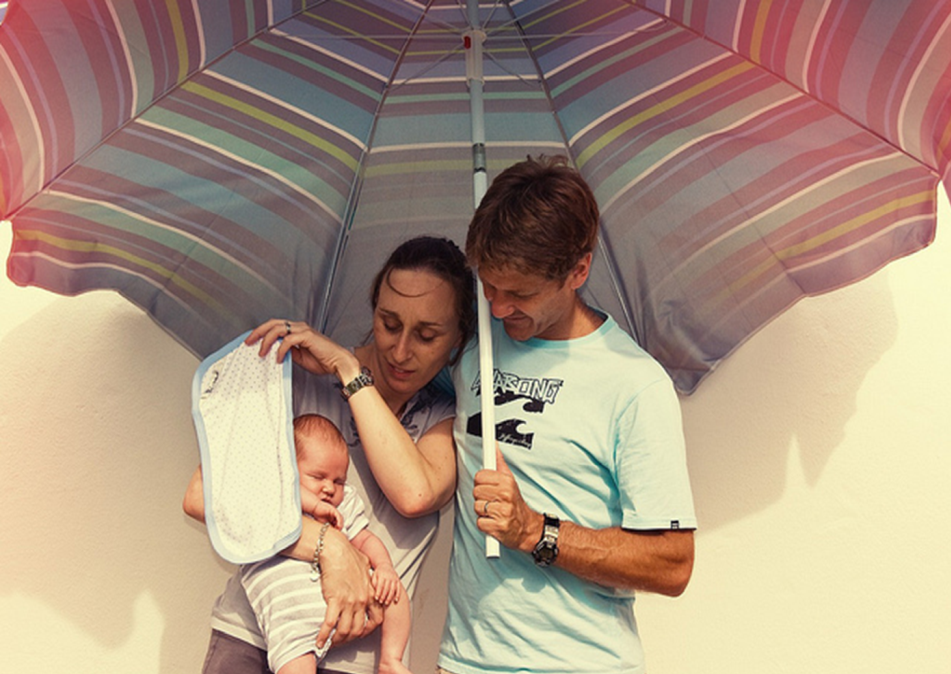}
  \end{minipage} \quad
  \begin{minipage}{.61\textwidth}
    \resizebox{\linewidth}{!}{
    \begin{tabular}{lll}
        \toprule
        \textbf{Mutation Type} & \textbf{Question} & \textbf{Answer} \\
        \toprule
            Original                    & Is the lady holding the baby?     & Yes \\
            Substitution (Negation)     & Is the lady not holding the baby? & No \\
            Substitution (Adversarial)  & Is the cat holding the baby?      & No \\
            \midrule
            Original & How many people are there? & Three \\
            Deletion (Masking) & How many [MASK] are there? & ``Number"\\
            \midrule
            Original & What is the color of the man's shirt? & Blue \\
            Substitution (Negation) & What is not the color of the man's shirt? & Magenta \\
            \midrule
            
            Deletion (Masking)          & Is the [MASK] holding the baby?   & Can't say \\ 
            \midrule
            Original & What color is the umbrella ? & Pink \\
            Deletion (Masking)          & What color is the [MASK]?   & ``color" \\
        \bottomrule
    \end{tabular}
    }
  \end{minipage}
  \captionof{table}{Examples of our question mutation.
The image is shown on the left, and the original question is in the first row of the table.
Examples of the two types of mutation are shown in the table.}
\label{tab:q_mut}
\end{figure*}

In order to train VQA models under the mutant paradigm, we need a mechanism to create mutant samples.
Mutations are transformations that act on semantic entities in either the image or the question, in ways that can reliably lead to a new answer.
For the question, semantic entities are words, while for images, semantic entities are objects.
It is important to note that our mutation process is automated and does not use the knowledge about the test set distribution in order to create new samples.
In this section, we delineate our automated generation process for both image and question-mutation.

    \subsection{Image Mutations}
    For image mutation, we first identify critical objects from the image that results in a change in the answer, and either remove instances of these objects (removal) or morph their color (substitution).
        
        \paragraph{Removing Object Instances:}
        Removing an instance of an object class can be either critical to the question (i.e. the answer to the question changes) or non-critical (i.e. answer is unchanged).
        If an object (or it's synonym or hypernym) is mentioned in the question, we deem it to be critical to the question, otherwise it is deemed non-critical.
        For each object with $M$ instances in the image, we randomly remove $m$ instances from the image s.t. $m \in \{0, \dots, M\}$ using polygon annotations from the COCO~\cite{lin2014microsoft} dataset.
        Thus for each image, we get multiple masked images, with pixels inside the instance bounding-box removed, as shown in Figure~\ref{fig:image_mutation}.
        These masked images are fed to a GAN-based inpainting network~\cite{yu2018generative} that makes the mutant image photo-realistic, and also prevents the model from getting cues from the shape of the mask. 
        In the case of numeric questions, if $m$ critical objects are removed, the answer to for the mutant image changes from $n$ to $n-m$.
        For yes-no questions, removal of all critical objects ($m=n$) will flip the answer from ``yes" to ``no", while removing $m<n$ critical objects will not.
        Note that $m=0$ corresponds to the original image and does not result in a change in the answer.
        
        \paragraph{Color Inversion:}
        For mutations that involve a change in color, we use samples with questions about the color of objects in the image, and change the color of critical objects by pixel-level color inversion in RGB-space.
        The true answer is replaced with the new color of the critical objects.
        To get objects with new colors, we do not use the knowledge about colors of objects in the world.
        In some cases, the new colors of the object may not correspond to real-world scenes, thus forcing the model to actually identifying colors, and not answer from language priors, such as ``bananas are yellow".

    \subsection{Question Mutations}
    We use three types of question mutations as shown in the example in Table~\ref{tab:q_mut}.
    We first identify the critical object and then apply template-based question operators similar to~\cite{gokhale2020vqa}.
    The first operator is negation for yes-no questions, which is achieved by a template based procedure that negates the question by adding a ``no" or ``not" before a verb, preposition or noun phrase.
    The second is the use of antonyms or adversarial object-words to substitute critical words.
    The third mutation masks words in the question and thus introduces ambiguity in the question.
    Questions for which the new answer cannot be deterministically identified are annotated with a broad category label such as \textit{color, location, fruit} instead of the exact answers such as \textit{red, library, apple} which the model cannot be expected to answer since some words have been masked or replaced with adversarial words.
    Yet, we want the model to be able to identify this broad category of answers even under partially occluded inputs.
    The answer remains unchanged for mutations with non-critical objects or words.
    

\begin{table}
    \centering
    \small
    \begin{tabular}{cc}
        \toprule
        \textbf{Mutation Category} & \textbf{Number of Samples} \\
        \midrule
        Object Removal              & 159,899   \\
        Color Change                & 30,759    \\
        Negation                    & 237,611   \\
        Adversarial Substitution    & 146,814   \\
        Word Masking                & 104,666   \\
        \bottomrule
    \end{tabular}

    \caption{Distribution of generated mutant samples by category of mutation}
    \label{tab:mutant_stats}
\end{table}    
    \subsection{Mutant Statistics:}
    We use the training set of VQA-CP-v2~\cite{vqa-cp} to generate mutant samples.
    For each original sample, we generate $1.5$ mutant samples on average, thus obtaining a total of 679k samples.
    Table~\ref{tab:mutant_stats} shows the distribution of our generated mutations with respect to the type of mutation.
    Addition of mutant samples does not change the distribution of samples per question-type.\footnote{More details about mutant samples are in Supp. material.}

\begin{table*}
	\small
	\begin{center}
		\resizebox{\linewidth}{!}{
			\begin{tabular}{@{}l  cccc c cccc c@{}}
				\toprule
				\multirow{2}{*}{Model}  & \multicolumn{4}{c}{VQA-CP v2 test (\%) $\uparrow$} & \hphantom & \multicolumn{4}{c}{VQA-v2 val (\%) $\uparrow$} & \multirow{2}{*}{Gap (\%)}\\
				\cmidrule{2-5} \cmidrule{7-10}
				& All & Yes/No & Num & Other && All & Yes/No & Num & Other & \\
				\toprule
				GVQA~\cite{agrawal2018don} & 31.30 & 57.99 & 13.68 & 22.14 && 48.24 & 72.03 & 31.17 & 34.65 & 16.94\\
				AReg~\cite{ramakrishnan2018overcoming} & 41.17 & 65.49 & 15.48 & 35.48 && 62.75 & 79.84 & 42.35 & 55.16 & 21.58\\
				RUBi~\cite{cadene2019rubi} & 47.11 & 68.65 & 20.28 & 43.18 && 63.10 & - & - & - & 14.05 \\ 
				SCR~\cite{wu2019self} & 48.47 & 70.41 & 10.42 & 47.29 && 62.30 & 77.40 & 40.90 & {56.50} & 13.83\\
			    LMH~\cite{clark2019don}  & 52.45 & 69.81 & 44.46 & 45.54 && 61.64 & 77.85 & 40.03 & 55.04 & 9.19\\
				CSS~\cite{chen2020counterfactual}  & 58.95 & 84.37 & 49.42 & 48.21 && 59.91 & 73.25 & 39.77 & 55.11 & {0.96}\\
				
				\midrule
				UpDn~\cite{anderson2018bottom}  & 39.74 & 42.27 & 11.93 & 46.05 && 63.48 & 81.18 & 42.14 & 55.66 & 23.74\\
				UpDn + Ours & 61.72 & 88.90 & 49.68 & 50.78 && 62.56 & 82.07 & 42.52 & 53.28 & 0.84\\
				\midrule
				LXMERT~\cite{tan2019lxmert} & 46.23 & 42.84 & 18.91 & 55.51 && \textbf{74.16} & \textbf{89.31} & \textbf{56.85} & \textbf{65.14} & 27.97\\
				LXMERT + Ours & \textbf{69.52} & \textbf{93.15} & \textbf{67.17} & \textbf{57.78} &&  \underline{70.24} & \underline{89.01} & \underline{54.21} & \underline{59.96} & \textbf{0.72} \\
				\bottomrule
			\end{tabular}
		} 
	\end{center}
	\caption{Accuracies on VQA-CP v2 test and VQA-v2 validation set, along with Percentage gap between overall accuracies on these two datasets. \textit{``Ours"} represents the final model with Answer Projection, Type Exposure and Pairwise Consistency.
	Overall best scores are \textbf{bold}, our best are \underline{underlined.}}
	\label{tab:results}
\end{table*}
\section{Experiments}
    \subsection{Setting}
    \paragraph{Datasets:}
    We train and evaluate our models on VQA-CP-v2.
    This is a natural choice for evaluating OOD generalization since VQA-CP is a non-i.i.d.~reorganization of the VQA dataset, and was created in order to evaluate VQA models in a setting where language priors cannot be relied upon for a correct prediction.
    This is because for every question type (65 types according to the question prefix), the prior distribution of answers is different in train and test splits of VQA-CP.
    We also train and evaluate our models on the VQA-v2~\cite{goyal2017making} validation set, and compare the gap between the imbalanced and non-i.i.d.~setting of VQA-CP against the balanced i.i.d.~setting of VQA.

    \paragraph{Hyperparameters:}
    All of our models are trained on two NVIDIA Tesla V100 16GB GPUs for $10$ epochs with batch size of $32$ and learning rate $1e\text{--}5$.
    Each epoch takes approximately three hours for UpDn and four hours for LXMERT.

    \subsection{Baseline Models}
    We compare our method with GVQA~\cite{agrawal2018don}, RUBI~\cite{cadene2019rubi}, SCR~\cite{wu2019self}, LMH~\cite{clark2019don}, CSS~\cite{chen2020counterfactual} as our baselines.
    Since most of these methods are built with UpDn~\cite{anderson2018bottom} as the backbone, we investigate the efficacy of UpDn under the mutant paradigm.
    On the other hand, LXMERT~\cite{tan2019lxmert} has emerged as a powerful transformer-based cross-modal feature extractor, and is pre-trained on tasks such as masked language modeling and cross-modality matching, inspired by BERT~\cite{devlin2018bert}.
    LXMERT is a top performing single-model on multiple vision-and-language tasks such as VQA, GQA~\cite{hudson2019gqa}, ViZWiz~\cite{bigham2010vizwiz}, and NLVR2~\cite{suhr2019corpus}.
    We therefore use is as a strong baseline for our experiments.
    LXMERT is representative of the recent trend towards using BERT-like pre-trained models~\cite{lu2019vilbert,su2019vl,li2020unicoder,chen2019uniter} and fine-tuning them on multiple downstream vision and language tasks.
    Note that we do not use ensemble models for our experiments and focus only on single-model baselines.

    \subsection{Results on VQA-CP-v2 and VQA-v2}
    Performance on two benchmarks VQA-CP-v2 and VQA-v2 is shown in Table~\ref{tab:results}.
    We compare existing models against UpDn and LXMERT incorporated into our Mutant method.
    For the VQA-CP benchmark, our method improves the performance of LXMERT by $23.29\%$, thus establishing a new state of the art on VQA-CP, beating the previous best by $10.57\%$.
    Our method shows improvements across all categories, with $8.78\%$ on the Yes-No category, $17.75\%$ on Number-based questions, and $9.57\%$ on other questions.
    We use negation as one of the question mutation operators on yes-no questions, but such questions are not present in the test set.
    However our model takes advantage of this mutation and improves substantially on yes-no questions.
    The Mutant method also consistently improves the performance of the UpDn model by $21.98\%$ overall.
    Note that baseline models AReg, RUBI, SCR, LMH, and CSS all modify UpDn by adding de-biasing techniques. 
    We show our de-biasing method improves on two SOTA models and outperforms all of the above baselines, unlike previous work which only modifies UpDn.
    This empirically shows Mutant to be model-agnostic.
    
    When trained and evaluated on the balanced i.i.d.~VQA-v2 dataset, our method achieves the best performance amongst methods designed specifically for OOD generalization, with an accuracy of $70.24\%$.
    This is closest among baselines to the SOTA established by LXMERT, which is trained explicitly for the balanced, i.i.d.~setting.
    To make this point clear, we report the \textit{gap} between the overall scores for VQA-CP and VQA-v2, following the protocol from~\citet{chen2020counterfactual} in Table~\ref{tab:results}.
    
    \paragraph{Results on VQA-v2 without re-training:\\}
    Additionally, we use our best model trained on VQA-CP and evaluate it on the VQA test standard set without re-training on VQA-v2 data.
    The objective here is to evaluate whether models trained on biased data (VQA-CP) and mutant data is able to generalize to VQA-v2 which uses an i.i.d. train-test split.
    This gives us an overall accuracy of $67.63\%$ comprising with $88.56\%$ on yes-no questions, $50.76\%$ on number-based questions, and $54.56\%$ on other questions.
    This is better than all existing VQA-CP models that are explicitly trained on VQA-v2 (reported in Table~\ref{tab:results}), and thus demonstrates the generalizability of our approach.

    \subsection{Analysis}
    \begin{table}[t]
    \centering
    \resizebox{\linewidth}{!}{
    \begin{tabular}{@{}p{1.75cm}>{\RaggedRight}p{3.45cm} p{0.7cm}p{0.7cm}p{0.7cm}p{0.7cm}}
        \toprule 
        \multirow{2}{*}{Model} & \multirow{2}{*}{Data} & \multicolumn{4}{c}{VQA-CP v2 test $\uparrow$ (\%)}\\
        \cmidrule{3-6}
        & & All & Yes/No & Num & Other \\
        \toprule
        UpDn    & VQA-CP            & 39.74 & 42.27 & 11.93 & 46.05 \\
        UpDn    & VQA-CP + Mutant   & 50.16 & 61.45 & 35.87 & 50.14 \\
        \hdashline
        \multicolumn{2}{c}{\textit{Increase in Accuracy}} & \textit{10.42} & \textit{19.18} & \textit{23.94} & \textit{4.09}\\
        \hdashline
        LXMERT  & VQA-CP            & 46.23 & 42.84 & 18.91 & 55.51 \\
        LXMERT  & VQA-CP + Mutant   & 59.69 & 73.19 & 32.85 & 59.29 \\
        \hdashline
        \multicolumn{2}{c}{\textit{Increase in Accuracy}} & \textit{13.46} & \textit{30.35} & \textit{13.94} & \textit{3.78}\\
        \midrule 
        LXM~+~Ours & VQA-CP + Img. Mut. & 64.85 & 85.68 & 66.44 & 53.80\\
        LXM~+~Ours & VQA-CP + Que. Mut. & 67.92 & 91.64 & 65.73 & 56.09\\
        LXM~+~Ours & VQA-CP + Both Mut. & \textbf{69.52} & \textbf{93.15} & \textbf{67.17} & \textbf{57.78} \\
        \bottomrule
    \end{tabular}
    }
    \caption{Top section: Comparison of UpDn and LXMERT when trained on VQA-CP and augmented with mutant samples, and the increase in accuracy due to mutant samples.
    Bottom section: Comparison of LXMERT when using image or text mutations, or both.}
    \label{tab:effect_data}
\end{table}
    \begin{table}[t]
    \centering
    \resizebox{\linewidth}{!}{
    \begin{tabular}{@{}lcccc@{}}
        \toprule 
        \multirow{2}{*}{Model} & \multicolumn{4}{c}{VQA-CP v2 test $\uparrow$ (\%)}\\
        \cmidrule{2-5}
        & All & Yes/No & Num & Other \\
        \toprule
        UpDn                    & 50.16 & 61.45 & 35.87 & 50.14 \\
        UpDn~+~AP             &  54.51 & 88.35 & 41.01 & 32.89\\
        UpDn~+~TE             &  56.32 & 80.56 & 46.14 & 46.41\\
        UpDn~+~AP~+~TE       &  55.76 & 90.25 & 43.78 & 41.40\\
        UpDn~+~AP~+~PW         &  57.54 & 91.59 & 49.17 & 41.93\\
        UpDn~+~TE~+~PW        &  60.32 & 86.10 & 50.23 & 49.58 \\
        UpDn~+~AP~+~TE~+~PW  &  61.72 & 88.90 & 49.68 & 50.78\\
        \midrule
        LXM                  & 59.69 & 73.19 & 32.85 & 59.29 \\
        LXM~+~AP            & 60.45 & 88.46 & 43.24 & 50.49 \\
        LXM~+~TE           & 63.36 & 77.10 & 46.50 & 61.27 \\
        LXM~+~AP~+~TE     & 64.73 & 85.34 & 47.23 & 58.71 \\
        LXM~+~AP~+~PW       & 67.14 & 90.49 & 65.52 & 55.34 \\
        LXM~+~TE~+~PW      & 64.17 & 94.71 & 35.19 & 48.80 \\
        LXM~+~AP~+~TE~+~PW & \textbf{69.52} & \textbf{93.15} & \textbf{67.17} & \textbf{57.78} \\
        
        \bottomrule
    \end{tabular}
    }
    \caption{Ablation study to investigate the effect of each component of our method: Answer Projection (AP), Type Exposure (TE), Pairwise Consistency (PW), and independent effect of image and question mutations.}
    \label{tab:ablation}
\end{table}

        \paragraph{Effect of Training with Mutant Samples:\\}
        In this analysis we measure the effect of augmenting the training data with mutant samples on UpDn and LXMERT without any architectural changes.
        The results are reported in Table~\ref{tab:effect_data}.
        Both models improve when exposed to the mutant samples, UpDn by $10.42\%$ and LXMERT by $13.46\%$.
        There is a markedly significant jump in performance for both models for the yes-no and number categories.
        UpDn especially benefits from Mutant samples in terms of the accuracy on numeric questions (a boost of $23.94\%$).
        
        We also compare our final model when trained only with image mutations and only with question mutations in Table~\ref{tab:effect_data}.
        While this is worse than training with both types of mutations, it can be seen that question mutations are better than image mutations in the case of yes-no and other questions, while image mutations are better on numeric questions.
        
        \paragraph{Ablation Study:\\}
        We conduct ablation studies to evaluate the efficacy of each component of our method, namely Answer Projection, Type Exposure and Pairwise Consistency, on both baselines, as shown in Table~\ref{tab:ablation}.
        Introduction of Answer Projection significantly improves yes-no performance, while Type Exposure improves performance on other questions.
        We also observe that the pairwise consistency loss significantly boosts performance on numeric questions and yes-no questions.
        Note that there is a minor difference between the original and the mutant sample, and the model needs to understand this difference, which in turn can enable the model to reason about the question and predict the new answer.
        For instance the pairwise consistency loss allows the model to learn the correlation between one missing object and a change in answer from ``two" to ``one" in Figure~\ref{fig:image_mutation}, resulting in an improvement in the counting ability of our VQA model.
        Similarly, the pairwise consistency allows the model to improve on yes-no questions for which the answer changes when a critical object is removed.
        
        \begin{table}
    \centering
    \resizebox{\linewidth}{!}{
    \begin{tabular}{@{}p{2.65cm}p{1.1cm}cccc@{}}
        \toprule 
        \multirow{2}{*}{Model} & \multirow{2}{*}{Method} & \multicolumn{4}{c}{VQA-CP v2 test $\uparrow$ (\%)}\\
        \cmidrule{3-6}
        & & All & Yes/No & Num & Other \\
        \toprule
        UpDn + Ours    & Base   & 61.72 & 88.90 & 49.68 & 50.78 \\
        UpDn + Ours    & LMH    & 55.38 & 90.99 & 39.74 & 40.99 \\
        \hdashline
        \multicolumn{2}{c}{\textit{Drop in Accuracy}} & \textit{6.34} & \textit{-2.09} & \textit{9.95} & \textit{9.80}\\
        \midrule
        LXMERT + Ours & Base    & 69.52 & 93.16 & 67.17 & 57.78 \\
        LXMERT + Ours & LMH     & 63.85 & 88.34 & 48.23 & 55.28\\
        \hdashline
        \multicolumn{2}{c}{\textit{Drop in Accuracy}} & \textit{5.67} & \textit{4.82} & \textit{18.86} & \textit{2.50} \\
        \bottomrule
    \end{tabular}
    }
    \caption{Effect of combining LMH de-biasing with the Mutant paradigm, measured as drop in accuracy (\%)}
    \label{tab:lmh}
\end{table}
        \paragraph{Effect of LMH Debiasing on Mutant:\\}
        We compare the results of our model when trained with or without the explicit de-biasing method LMH~\cite{clark2019don}.
        LMH is an ensemble-based method trained for \textit{avoiding} dataset biases, and is the most effective among all de-biasing strategies developed for the VQA-CP challenge.
        LMH implements a learned mixing strategy, by using the main model in combination with a bias-only model trained only with the question, without the image.
        The learned mixing strategy uses the bias-only model to remove biases from the main model.
        It can be seen from Table~\ref{tab:lmh} that LMH leads to a drop in performance when used in combination with Mutant.
        This is potentially because in the process of debiasing, LMH ends up attenuating positive bias introduced by Mutant that is useful for generalization.
        \citet{kervadec2020roses} have concurrently shown that de-biasing methods such as LMH indeed result in a decrease in performance on out-of-distribution (OOD) test samples in the GQA~\cite{hudson2019gqa} dataset, mirroring our analysis on VQA-CP shown in Table~\ref{tab:lmh}.

\section{Related Work} 

    \paragraph{De-biasing of VQA datasets:}
    The VQA-v1 dataset~\cite{antol2015vqa} contained imbalances and language priors between question- answer pairs.
    This was mitigated by VQA-v2~\cite{goyal2017making} which balanced the data by collecting complementary images such that each question was associated with two images leading to two different answers.
    Identifying that the distribution of answers in the VQA dataset led models to learn superficial correlations,~\citet{vqa-cp} proposed the VQA-CP dataset by re-organizing the train and test splits such that the the distribution of answers per question-type was significantly different for each split.

    \paragraph{Robustness in VQA:}
    Ongoing efforts seek to build robust VQA models for VQA for various aspects of robustness.
    \citet{shah2019cycle} propose a model that uses cycle-consistency to not only answer the question, but also generate a complimentary question with the same answer, in order to increase the linguistic diversity of questions. 
    In constrast, our work generates questions with a different answer.
    \citet{selvaraju2020squinting} provide a dataset which contains perception-related sub-questions for each VQA question.
    Antonym-consistency has been tackled in~\citet{ray2019sunny}.
    Inspired by invariant risk minimization~\cite{arjovsky2019invariant} which links out-of-distribution generalization to invariance and causality,~\citet{teney2020unshuffling} provide a method to identify invariant correlations in the training set and train models to ignore spurious correlations.
    \citet{asai2020logic,gokhale2020vqa} explore robustness to logical transformation of questions using first-order logic connectives \textit{and} ($\wedge$), \textit{or} ($\vee$), \textit{not} ($\neg$).
    Removal of bias has been a focus of ~\citet{ramakrishnan2018overcoming,clark2019don} for the VQA-CP task.
    We distinguish our work from these by amplifying positive bias and attenuating negative bias.
    
    \paragraph{Data Augmentation:}
    It is important to note that the above work on data de-biasing and robust models focuses on the language priors in VQA, but not much attention has been given to visual priors.
    Within the last year, there has been interest in augmenting VQA training data with counterfactual images~\cite{agarwal2020towards,chen2020counterfactual}. Independently,~\citet{teney2020learning} have also demonstrated that counterfactual images obtained via minimal editing such as masking or inpainting can lead to improved OOD generalization of VQA models, when trained with a pairwise gradient-based regularization.
    Self-supervised data augmentation has been explored in recent work~\cite{lewis-etal-2019-unsupervised,fabbri2020template,banerjee2020self} in the domain of text-based question answering.
    The mutant paradigm presented in this work is one of the first enable the generation of VQA samples that result in different answers, coupled with a novel architecture and a consistency loss between original and mutant samples as a training objective.

    \paragraph{Answer Embeddings:}
    In one of the early works on VQA,~\citet{teney2016zero} use a combination of image and question representations and answer embeddings to predict the final answer. 
    \citet{hu2018learning} learn two embedding functions that transform image-question pair and answers to a shared latent space.
     Our method is different from this since we use a combination of classification and NCE Loss on the projection of answer vectors, as opposed to a single training objective.
    This means that although the predicted answer is obtained as the most probable answer from a set of candidate answers, the NCE Loss in the answer-space embeds the notion of semantic similarity between the answer.
    Our Type Exposure model is in principal similar to~\citet{kafle2016answer} who use the predicted answer-type probabilities in a Bayesian framework, while we use it as an additional constraint, i.e. as a regularization for a maximum likelihood objective.

\section{Discussion and Conclusion}
In this paper, we present a method that uses input mutations to train VQA models with the goal of Out-of-Distribution generalization.
Our novel answer projection module trained for minimizing distance between answer and input projections complements the canonical VQA classification task.
Our Type Exposure model allows our network to consider all valid answers per question type as equally probable answer candidates, thus moving away from the negative question-answer linguistic priors.
Coupled with pairwise consistency, these modules achieve a new state-of-the-art accuracy on the VQA-CP-v2 dataset and reduce the gap between model performance on VQA-v2 data.

We differentiate our work from methods using random adversarial perturbations for robust learning~\cite{madry2018towards}.
Instead we view input mutations as \textit{structured perturbations} which lead to a semantic change in the input space and a deterministic change in the output space.
We envision that the concept of input mutations can be extended to other vision and language tasks for robustness.
Concurrent work in the domain of image classification shows that carefully designed perturbations or manipulations of the input can benefit generalization and lead to performance improvements~\cite{chen2020simple,hendrycks2019augmix}.
While perception is a cornerstone of understanding, the ability to imagine changes in the scene or language query, and predict outputs for that \textit{imagined} input allows models to supplement ``what" decision making (based on observed inputs) with ``what if" decision making (based on imagined inputs).
The Mutant paradigm is an effort towards ``what if" decision making. Code is available \href{https://github.com/ASU-Active-Perception-Group}{here}.

\section*{Acknowledgements}
The authors acknowledge support from the NSF Robust Intelligence Program project \#1816039, the DARPA KAIROS program (LESTAT project), the DARPA SAIL-ON program, and ONR award N00014-20-1-2332.

\bibliography{emnlp2020}

\begin{thebibliography}{51}
\expandafter\ifx\csname natexlab\endcsname\relax\def\natexlab#1{#1}\fi

\bibitem[{Agarwal et~al.(2020)Agarwal, Shetty, and Fritz}]{agarwal2020towards}
Vedika Agarwal, Rakshith Shetty, and Mario Fritz. 2020.
\newblock Towards causal vqa: Revealing and reducing spurious correlations by
  invariant and covariant semantic editing.
\newblock In \emph{Proceedings of the IEEE/CVF Conference on Computer Vision
  and Pattern Recognition}, pages 9690--9698.

\bibitem[{Agrawal et~al.(2018{\natexlab{a}})Agrawal, Batra, Parikh, and
  Kembhavi}]{vqa-cp}
Aishwarya Agrawal, Dhruv Batra, Devi Parikh, and Aniruddha Kembhavi.
  2018{\natexlab{a}}.
\newblock Don't just assume; look and answer: Overcoming priors for visual
  question answering.
\newblock In \emph{IEEE Conference on Computer Vision and Pattern Recognition
  (CVPR)}.

\bibitem[{Agrawal et~al.(2018{\natexlab{b}})Agrawal, Batra, Parikh, and
  Kembhavi}]{agrawal2018don}
Aishwarya Agrawal, Dhruv Batra, Devi Parikh, and Aniruddha Kembhavi.
  2018{\natexlab{b}}.
\newblock Don't just assume; look and answer: Overcoming priors for visual
  question answering.
\newblock In \emph{CVPR}.

\bibitem[{Anderson et~al.(2018)Anderson, He, Buehler, Teney, Johnson, Gould,
  and Zhang}]{anderson2018bottom}
Peter Anderson, Xiaodong He, Chris Buehler, Damien Teney, Mark Johnson, Stephen
  Gould, and Lei Zhang. 2018.
\newblock Bottom-up and top-down attention for image captioning and visual
  question answering.
\newblock In \emph{Proceedings of the IEEE conference on computer vision and
  pattern recognition}, pages 6077--6086.

\bibitem[{Antol et~al.(2015)Antol, Agrawal, Lu, Mitchell, Batra,
  Lawrence~Zitnick, and Parikh}]{antol2015vqa}
Stanislaw Antol, Aishwarya Agrawal, Jiasen Lu, Margaret Mitchell, Dhruv Batra,
  C~Lawrence~Zitnick, and Devi Parikh. 2015.
\newblock Vqa: Visual question answering.
\newblock In \emph{Proceedings of the IEEE international conference on computer
  vision}, pages 2425--2433.

\bibitem[{Arjovsky et~al.(2019)Arjovsky, Bottou, Gulrajani, and
  Lopez-Paz}]{arjovsky2019invariant}
Martin Arjovsky, L{\'e}on Bottou, Ishaan Gulrajani, and David Lopez-Paz. 2019.
\newblock Invariant risk minimization.
\newblock \emph{arXiv preprint arXiv:1907.02893}.

\bibitem[{Asai and Hajishirzi(2020)}]{asai2020logic}
Akari Asai and Hannaneh Hajishirzi. 2020.
\newblock Logic-guided data augmentation and regularization for consistent
  question answering.
\newblock In \emph{ACL}.

\bibitem[{Banerjee and Baral(2020)}]{banerjee2020self}
Pratyay Banerjee and Chitta Baral. 2020.
\newblock Self-supervised knowledge triplet learning for zero-shot question
  answering.
\newblock In \emph{EMNLP}.

\bibitem[{Bigham et~al.(2010)Bigham, Jayant, Ji, Little, Miller, Miller,
  Miller, Tatarowicz, White, White et~al.}]{bigham2010vizwiz}
Jeffrey~P Bigham, Chandrika Jayant, Hanjie Ji, Greg Little, Andrew Miller,
  Robert~C Miller, Robin Miller, Aubrey Tatarowicz, Brandyn White, Samual
  White, et~al. 2010.
\newblock Vizwiz: nearly real-time answers to visual questions.
\newblock In \emph{Proceedings of the 23nd annual ACM symposium on User
  interface software and technology}, pages 333--342.

\bibitem[{Cadene et~al.(2019)Cadene, Dancette, Ben-younes, Cord, and
  Parikh}]{cadene2019rubi}
Remi Cadene, Corentin Dancette, Hedi Ben-younes, Matthieu Cord, and Devi
  Parikh. 2019.
\newblock Rubi: Reducing unimodal biases in visual question answering.
\newblock In \emph{NeurIPS}.

\bibitem[{Chen et~al.(2020{\natexlab{a}})Chen, Yan, Xiao, Zhang, Pu, and
  Zhuang}]{chen2020counterfactual}
Long Chen, Xin Yan, Jun Xiao, Hanwang Zhang, Shiliang Pu, and Yueting Zhuang.
  2020{\natexlab{a}}.
\newblock Counterfactual samples synthesizing for robust visual question
  answering.
\newblock In \emph{Proceedings of the IEEE/CVF Conference on Computer Vision
  and Pattern Recognition}, pages 10800--10809.

\bibitem[{Chen et~al.(2020{\natexlab{b}})Chen, Kornblith, Norouzi, and
  Hinton}]{chen2020simple}
Ting Chen, Simon Kornblith, Mohammad Norouzi, and Geoffrey Hinton.
  2020{\natexlab{b}}.
\newblock A simple framework for contrastive learning of visual
  representations.
\newblock In \emph{International Conference on Machine Learning}.

\bibitem[{Chen et~al.(2019)Chen, Li, Yu, Kholy, Ahmed, Gan, Cheng, and
  Liu}]{chen2019uniter}
Yen-Chun Chen, Linjie Li, Licheng Yu, Ahmed~El Kholy, Faisal Ahmed, Zhe Gan,
  Yu~Cheng, and Jingjing Liu. 2019.
\newblock Uniter: Learning universal image-text representations.
\newblock In \emph{European Conference on Computer Vision}.

\bibitem[{Clark et~al.(2019)Clark, Yatskar, and Zettlemoyer}]{clark2019don}
Christopher Clark, Mark Yatskar, and Luke Zettlemoyer. 2019.
\newblock Don’t take the easy way out: Ensemble based methods for avoiding
  known dataset biases.
\newblock In \emph{Proceedings of the 2019 Conference on Empirical Methods in
  Natural Language Processing and the 9th International Joint Conference on
  Natural Language Processing (EMNLP-IJCNLP)}, pages 4060--4073.

\bibitem[{Deng et~al.(2009)Deng, Dong, Socher, Li, Li, and
  Fei-Fei}]{deng2009imagenet}
Jia Deng, Wei Dong, Richard Socher, Li-Jia Li, Kai Li, and Li~Fei-Fei. 2009.
\newblock Imagenet: A large-scale hierarchical image database.
\newblock In \emph{2009 IEEE conference on computer vision and pattern
  recognition}, pages 248--255. Ieee.

\bibitem[{Devlin et~al.(2019)Devlin, Chang, Lee, and
  Toutanova}]{devlin2018bert}
Jacob Devlin, Ming-Wei Chang, Kenton Lee, and Kristina Toutanova. 2019.
\newblock Bert: Pre-training of deep bidirectional transformers for language
  understanding.
\newblock In \emph{NAACL-HLT (1)}.

\bibitem[{Fabbri et~al.(2020)Fabbri, Ng, Wang, Nallapati, and
  Xiang}]{fabbri2020template}
Alexander~R Fabbri, Patrick Ng, Zhiguo Wang, Ramesh Nallapati, and Bing Xiang.
  2020.
\newblock Template-based question generation from retrieved sentences for
  improved unsupervised question answering.
\newblock In \emph{ACL}.

\bibitem[{Gokhale et~al.(2020)Gokhale, Banerjee, Baral, and
  Yang}]{gokhale2020vqa}
Tejas Gokhale, Pratyay Banerjee, Chitta Baral, and Yezhou Yang. 2020.
\newblock Vqa-lol: Visual question answering under the lens of logic.
\newblock In \emph{European conference on computer vision}. Springer.

\bibitem[{Goyal et~al.(2017)Goyal, Khot, Summers-Stay, Batra, and
  Parikh}]{goyal2017making}
Yash Goyal, Tejas Khot, Douglas Summers-Stay, Dhruv Batra, and Devi Parikh.
  2017.
\newblock Making the v in vqa matter: Elevating the role of image understanding
  in visual question answering.
\newblock In \emph{Proceedings of the IEEE Conference on Computer Vision and
  Pattern Recognition}, pages 6904--6913.

\bibitem[{Gutmann and Hyv{\"a}rinen(2010)}]{gutmann2010noise}
Michael Gutmann and Aapo Hyv{\"a}rinen. 2010.
\newblock Noise-contrastive estimation: A new estimation principle for
  unnormalized statistical models.
\newblock In \emph{Proceedings of the Thirteenth International Conference on
  Artificial Intelligence and Statistics}, pages 297--304.

\bibitem[{Hendrycks et~al.(2019)Hendrycks, Mu, Cubuk, Zoph, Gilmer, and
  Lakshminarayanan}]{hendrycks2019augmix}
Dan Hendrycks, Norman Mu, Ekin~Dogus Cubuk, Barret Zoph, Justin Gilmer, and
  Balaji Lakshminarayanan. 2019.
\newblock Augmix: A simple data processing method to improve robustness and
  uncertainty.
\newblock In \emph{International Conference on Learning Representations}.

\bibitem[{Honnibal and Montani(2017)}]{honnibal2017spacy}
Matthew Honnibal and Ines Montani. 2017.
\newblock spacy 2: Natural language understanding with bloom embeddings,
  convolutional neural networks and incremental parsing.
\newblock \emph{To appear}, 7(1).

\bibitem[{Hu et~al.(2018)Hu, Chao, and Sha}]{hu2018learning}
Hexiang Hu, Wei-Lun Chao, and Fei Sha. 2018.
\newblock Learning answer embeddings for visual question answering.
\newblock In \emph{Proceedings of the IEEE Conference on Computer Vision and
  Pattern Recognition}, pages 5428--5436.

\bibitem[{Hudson and Manning(2019)}]{hudson2019gqa}
Drew~A Hudson and Christopher~D Manning. 2019.
\newblock Gqa: A new dataset for real-world visual reasoning and compositional
  question answering.
\newblock In \emph{Proceedings of the IEEE Conference on Computer Vision and
  Pattern Recognition}, pages 6700--6709.

\bibitem[{Jascob(v0.2.1 (February 22, 2020))}]{lemminflect}
Brad Jascob. v0.2.1 (February 22, 2020).
\newblock Lemminflect. a python module for english word lemmatization and
  inflection.
\newblock \url{https://github.com/bjascob/LemmInflect}.

\bibitem[{Kafle and Kanan(2016)}]{kafle2016answer}
Kushal Kafle and Christopher Kanan. 2016.
\newblock Answer-type prediction for visual question answering.
\newblock In \emph{Proceedings of the IEEE Conference on Computer Vision and
  Pattern Recognition}, pages 4976--4984.

\bibitem[{Kaushik and Lipton(2018)}]{kaushik2018much}
Divyansh Kaushik and Zachary~C Lipton. 2018.
\newblock How much reading does reading comprehension require? a critical
  investigation of popular benchmarks.
\newblock In \emph{Proceedings of the 2018 Conference on Empirical Methods in
  Natural Language Processing}, pages 5010--5015.

\bibitem[{Kervadec et~al.(2020)Kervadec, Antipov, Baccouche, and
  Wolf}]{kervadec2020roses}
Corentin Kervadec, Grigory Antipov, Moez Baccouche, and Christian Wolf. 2020.
\newblock Roses are red, violets are blue... but should vqa expect them to?
\newblock \emph{arXiv preprint arXiv:2006.05121}.

\bibitem[{Lewis et~al.(2019)Lewis, Denoyer, and
  Riedel}]{lewis-etal-2019-unsupervised}
Patrick Lewis, Ludovic Denoyer, and Sebastian Riedel. 2019.
\newblock \href {https://doi.org/10.18653/v1/P19-1484} {Unsupervised question
  answering by cloze translation}.
\newblock In \emph{Proceedings of the 57th Annual Meeting of the Association
  for Computational Linguistics}, pages 4896--4910, Florence, Italy.
  Association for Computational Linguistics.

\bibitem[{Li et~al.(2020)Li, Duan, Fang, Gong, Jiang, and
  Zhou}]{li2020unicoder}
Gen Li, Nan Duan, Yuejian Fang, Ming Gong, Daxin Jiang, and Ming Zhou. 2020.
\newblock Unicoder-vl: A universal encoder for vision and language by
  cross-modal pre-training.
\newblock In \emph{AAAI}, pages 11336--11344.

\bibitem[{Lin et~al.(2014)Lin, Maire, Belongie, Hays, Perona, Ramanan,
  Doll{\'a}r, and Zitnick}]{lin2014microsoft}
Tsung-Yi Lin, Michael Maire, Serge Belongie, James Hays, Pietro Perona, Deva
  Ramanan, Piotr Doll{\'a}r, and C~Lawrence Zitnick. 2014.
\newblock Microsoft coco: Common objects in context.
\newblock In \emph{European conference on computer vision}, pages 740--755.
  Springer.

\bibitem[{Lloyd(1982)}]{lloyd1982least}
Stuart Lloyd. 1982.
\newblock Least squares quantization in pcm.
\newblock \emph{IEEE transactions on information theory}, 28(2):129--137.

\bibitem[{Lu et~al.(2019)Lu, Batra, Parikh, and Lee}]{lu2019vilbert}
Jiasen Lu, Dhruv Batra, Devi Parikh, and Stefan Lee. 2019.
\newblock Vilbert: Pretraining task-agnostic visiolinguistic representations
  for vision-and-language tasks.
\newblock In \emph{Advances in Neural Information Processing Systems}, pages
  13--23.

\bibitem[{Madry et~al.(2018)Madry, Makelov, Schmidt, Tsipras, and
  Vladu}]{madry2018towards}
Aleksander Madry, Aleksandar Makelov, Ludwig Schmidt, Dimitris Tsipras, and
  Adrian Vladu. 2018.
\newblock \href {https://openreview.net/forum?id=rJzIBfZAb} {Towards deep
  learning models resistant to adversarial attacks}.
\newblock In \emph{International Conference on Learning Representations}.

\bibitem[{McCoy et~al.(2019)McCoy, Pavlick, and Linzen}]{mccoy2019right}
Tom McCoy, Ellie Pavlick, and Tal Linzen. 2019.
\newblock Right for the wrong reasons: Diagnosing syntactic heuristics in
  natural language inference.
\newblock In \emph{Proceedings of the 57th Annual Meeting of the Association
  for Computational Linguistics}, pages 3428--3448.

\bibitem[{Mitchell(1980)}]{mitchell1980need}
Tom~M Mitchell. 1980.
\newblock \emph{The need for biases in learning generalizations}.
\newblock Department of Computer Science, Laboratory for Computer Science
  Research~….

\bibitem[{Pennington et~al.(2014)Pennington, Socher, and
  Manning}]{pennington2014glove}
Jeffrey Pennington, Richard Socher, and Christopher~D Manning. 2014.
\newblock Glove: Global vectors for word representation.
\newblock In \emph{Proceedings of the 2014 conference on empirical methods in
  natural language processing (EMNLP)}, pages 1532--1543.

\bibitem[{Poliak et~al.(2018)Poliak, Naradowsky, Haldar, Rudinger, and
  Van~Durme}]{poliak2018hypothesis}
Adam Poliak, Jason Naradowsky, Aparajita Haldar, Rachel Rudinger, and Benjamin
  Van~Durme. 2018.
\newblock Hypothesis only baselines in natural language inference.
\newblock In \emph{Proceedings of the Seventh Joint Conference on Lexical and
  Computational Semantics}, pages 180--191.

\bibitem[{Ramakrishnan et~al.(2018)Ramakrishnan, Agrawal, and
  Lee}]{ramakrishnan2018overcoming}
Sainandan Ramakrishnan, Aishwarya Agrawal, and Stefan Lee. 2018.
\newblock Overcoming language priors in visual question answering with
  adversarial regularization.
\newblock In \emph{Advances in Neural Information Processing Systems}, pages
  1541--1551.

\bibitem[{Ray et~al.(2019)Ray, Sikka, Divakaran, Lee, and
  Burachas}]{ray2019sunny}
Arijit Ray, Karan Sikka, Ajay Divakaran, Stefan Lee, and Giedrius Burachas.
  2019.
\newblock Sunny and dark outside?! improving answer consistency in vqa through
  entailed question generation.
\newblock In \emph{Proceedings of the 2019 Conference on Empirical Methods in
  Natural Language Processing and the 9th International Joint Conference on
  Natural Language Processing (EMNLP-IJCNLP)}, pages 5863--5868.

\bibitem[{Selvaraju et~al.(2020)Selvaraju, Tendulkar, Parikh, Horvitz, Ribeiro,
  Nushi, and Kamar}]{selvaraju2020squinting}
Ramprasaath~R. Selvaraju, Purva Tendulkar, Devi Parikh, Eric Horvitz, Marco
  Ribeiro, Besmira Nushi, and Ece Kamar. 2020.
\newblock Squinting at vqa models: Interrogating vqa models with sub-questions.
\newblock In \emph{IEEE Conference on Computer Vision and Pattern Recognition
  (CVPR)}.

\bibitem[{Shah et~al.(2019)Shah, Chen, Rohrbach, and Parikh}]{shah2019cycle}
Meet Shah, Xinlei Chen, Marcus Rohrbach, and Devi Parikh. 2019.
\newblock Cycle-consistency for robust visual question answering.
\newblock In \emph{Proceedings of the IEEE Conference on Computer Vision and
  Pattern Recognition}, pages 6649--6658.

\bibitem[{Su et~al.(2019)Su, Zhu, Cao, Li, Lu, Wei, and Dai}]{su2019vl}
Weijie Su, Xizhou Zhu, Yue Cao, Bin Li, Lewei Lu, Furu Wei, and Jifeng Dai.
  2019.
\newblock Vl-bert: Pre-training of generic visual-linguistic representations.
\newblock In \emph{International Conference on Learning Representations}.

\bibitem[{Suhr et~al.(2019)Suhr, Zhou, Zhang, Zhang, Bai, and
  Artzi}]{suhr2019corpus}
Alane Suhr, Stephanie Zhou, Ally Zhang, Iris Zhang, Huajun Bai, and Yoav Artzi.
  2019.
\newblock A corpus for reasoning about natural language grounded in
  photographs.
\newblock In \emph{Proceedings of the Annual Meeting of the Association for
  Computational Linguistics}.

\bibitem[{Tan and Bansal(2019)}]{tan2019lxmert}
Hao Tan and Mohit Bansal. 2019.
\newblock Lxmert: Learning cross-modality encoder representations from
  transformers.
\newblock In \emph{Proceedings of the 2019 Conference on Empirical Methods in
  Natural Language Processing and the 9th International Joint Conference on
  Natural Language Processing (EMNLP-IJCNLP)}, pages 5103--5114.

\bibitem[{Teney et~al.(2020{\natexlab{a}})Teney, Abbasnedjad, and
  Hengel}]{teney2020learning}
Damien Teney, Ehsan Abbasnedjad, and Anton van~den Hengel. 2020{\natexlab{a}}.
\newblock Learning what makes a difference from counterfactual examples and
  gradient supervision.
\newblock In \emph{European conference on computer vision}. Springer.

\bibitem[{Teney et~al.(2020{\natexlab{b}})Teney, Abbasnejad, and
  Hengel}]{teney2020unshuffling}
Damien Teney, Ehsan Abbasnejad, and Anton van~den Hengel. 2020{\natexlab{b}}.
\newblock Unshuffling data for improved generalization.
\newblock \emph{arXiv preprint arXiv:2002.11894}.

\bibitem[{Teney and Hengel(2016)}]{teney2016zero}
Damien Teney and Anton van~den Hengel. 2016.
\newblock Zero-shot visual question answering.
\newblock \emph{arXiv preprint arXiv:1611.05546}.

\bibitem[{Willing et~al.(2011)Willing, Russell, and
  Finlay}]{willing2011shifting}
Benjamin~P Willing, Shannon~L Russell, and B~Brett Finlay. 2011.
\newblock Shifting the balance: antibiotic effects on host--microbiota
  mutualism.
\newblock \emph{Nature Reviews Microbiology}, 9(4):233--243.

\bibitem[{Wu and Mooney(2019)}]{wu2019self}
Jialin Wu and Raymond~J Mooney. 2019.
\newblock Self-critical reasoning for robust visual question answering.
\newblock In \emph{NeurIPS}.

\bibitem[{Yu et~al.(2018)Yu, Lin, Yang, Shen, Lu, and Huang}]{yu2018generative}
Jiahui Yu, Zhe Lin, Jimei Yang, Xiaohui Shen, Xin Lu, and Thomas~S Huang. 2018.
\newblock Generative image inpainting with contextual attention.
\newblock In \emph{Proceedings of the IEEE conference on computer vision and
  pattern recognition}, pages 5505--5514.

\end{thebibliography}
\bibliographystyle{acl_natbib}

\clearpage
\appendix
\section*{Appendix}

\section{Datasets}
    \subsection{VQA-CP}
    VQA-CP (Visual Question Answering under Changing Priors)~\cite{vqa-cp} is a re-organization of the VQA dataset~\cite{antol2015vqa,goyal2017making}.
    The aim of VQA-CP is to have a different distribution of answers per question type is different in test and train splits.
    There are 65 question types based on the prefix of the questions such as \textit{``how many", ``what color", ``what sport", ``is there", "what is the", ``which"}.
    In VQA-v2, samples are drawn at randomly and independently and assigned either to train or test, thus resulting in the same distribution for both splits.
    \begin{equation*}
        P^{VQA}_{train}(A | Q, I) = P^{VQA}_{test}(A | Q, I).
    \end{equation*}
    
    In VQA-CP however, samples are assigned using a greedy re-splitting algorithm,  either to train or test, in a way that makes sure that questions with the same type an same answer are not shared by train and test.
    It is important to note that there is no leakage between train and test splits compared to the original VQA splits.
    \begin{equation*}
        P^{VQA-CP}_{train}(A | Q, I) \neq P^{VQA-CP}_{test}(A | Q, I).
    \end{equation*}
    
    The train set for VQA-CP-v2 contains 121k images, 245k questions and 2.5M answers, while the test set contains 98k images, 220k questions and 2.2M answers.
    
    \subsection{COCO}
    The source of images in both VQA and VQA-CP is the MS-COCO dataset~\cite{lin2014microsoft}.
    COCO contains natural images representing complex, real-world scenes containing common objects of 91 categories such as \textit{``person", ``chair", ``fork", ``horse", ``sports-ball"}, etc.
    For each image, COCO provides 5 captions along with bounding boxes and polygon annotations for each object instance in the image.

\begin{figure}[t]
    \centering
    \includegraphics[width=\linewidth]{ 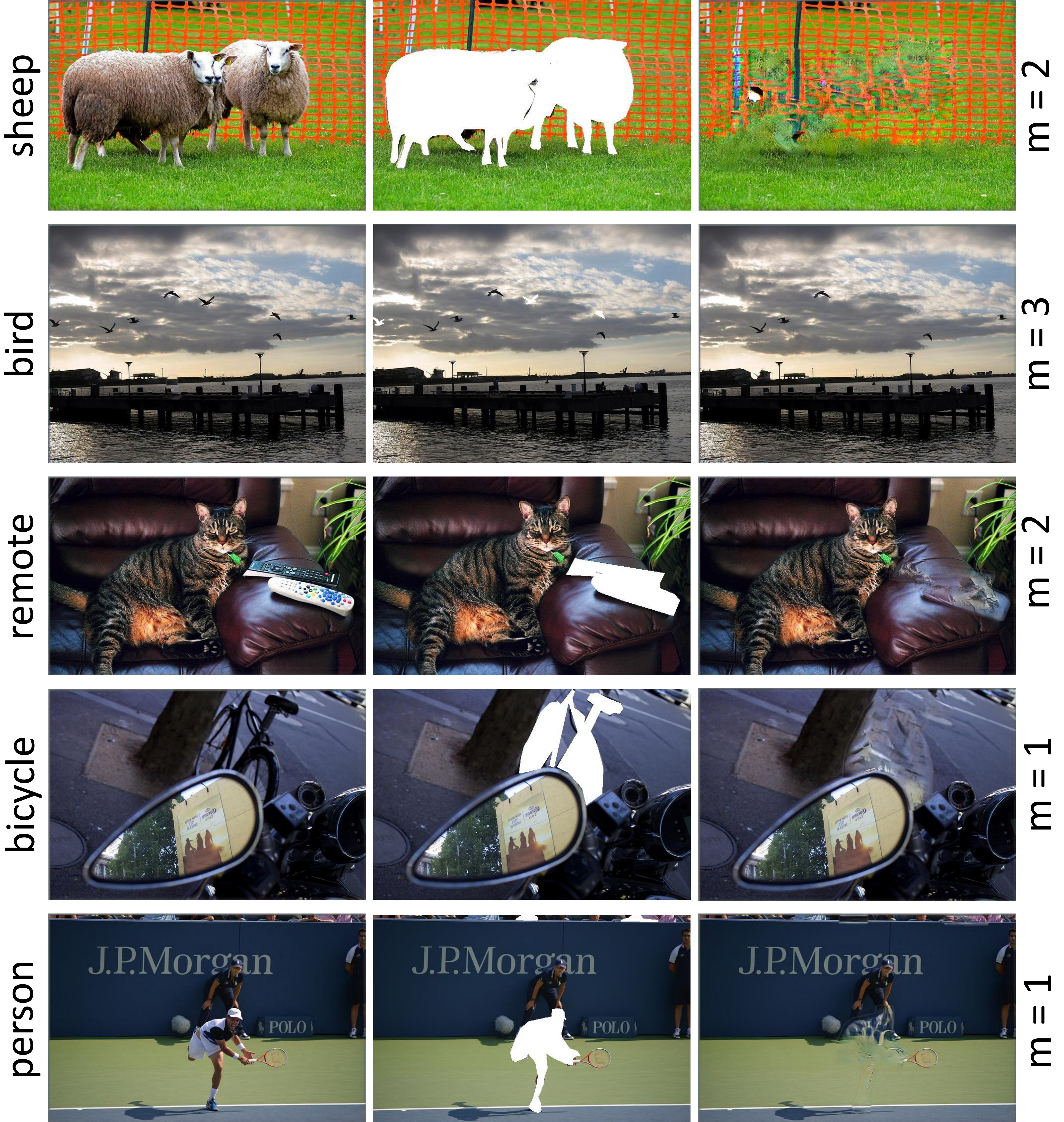}
    \caption{Illustration of COCO bounding box and polygon annotations for $m$ instances of an object, and the inpainting results after removal}
   
    \label{fig:bbox}
\end{figure}

\begin{figure}[t]
    \centering
    \includegraphics[width=\linewidth]{ 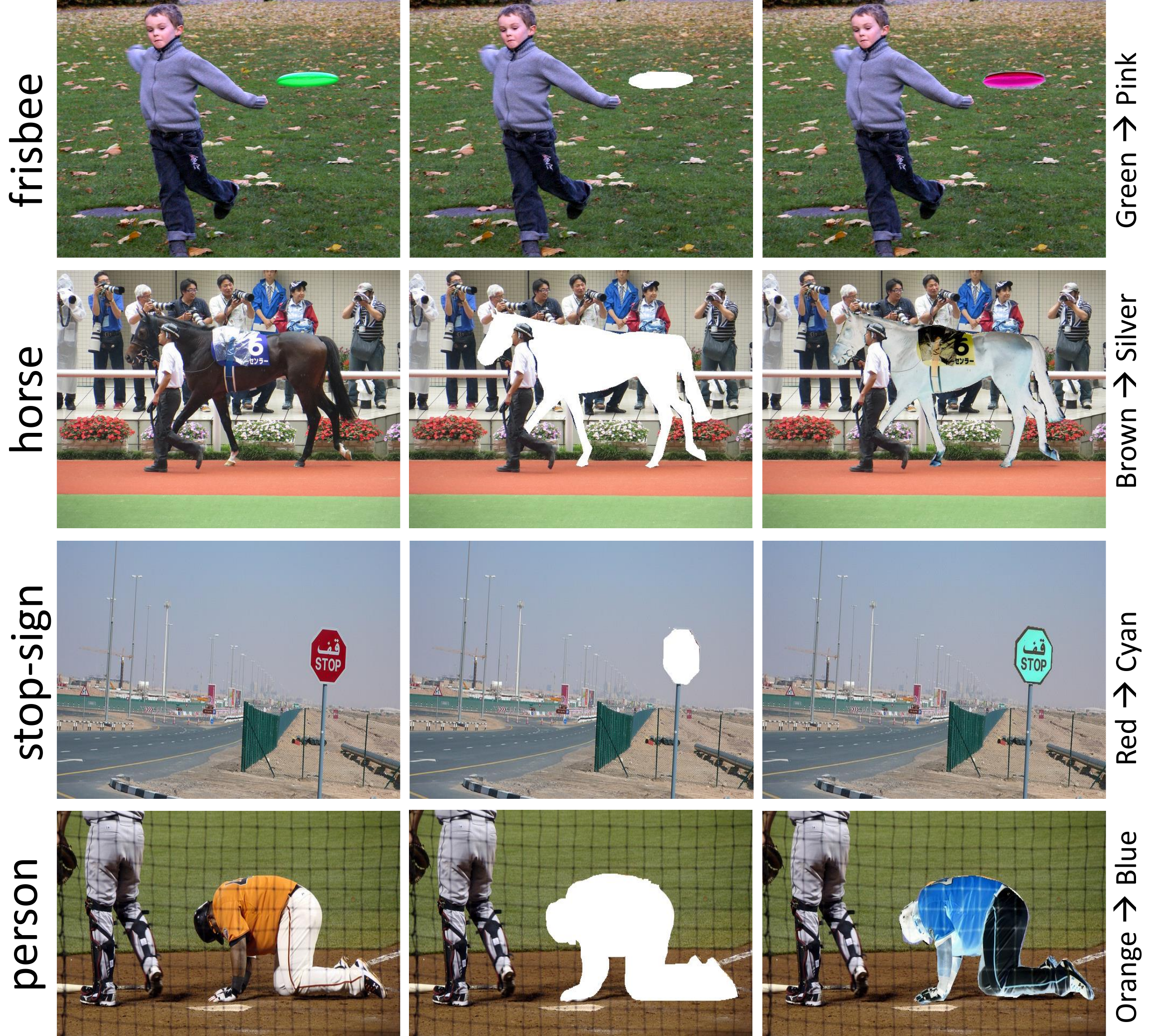}
    \caption{Illustration of color inversion procedure}
    \label{fig:col}
\end{figure}

\section{Image Mutant Generation Process}
    
In this section we provide additional details about our process for generating mutant samples from original question-image-answer triplets (Q-I-A) in the VQA-CP dataset.
For all linguistic operations we use a combination of SpaCy~\cite{honnibal2017spacy} and the LemmInflect library~\cite{lemminflect}
for lemmatization and inflection.

    \subsection{Selection of Objects}
    For each VQA sample, a list of words $W$ is created, which contains words from the ground-truth answers and the question.
    All nouns in $W$ are converted to their singular form.
    For yes-no questions, numeric questions, and questions about colors of objects, a list of objects $O$ is obtained from COCO.
    Background and crowd objects are filtered out from $O$.
    From $O$ critical objects $O_C$ and and non-critical objects $O_{NC}$ are obtained.
    Critical objects are those objects in the image that when manipulated or removed, may change the answer to the question being asked.
    For this we follow a simple heuristic that states that if an object-word or it's synonym or hyponym is present in $W$, then it is a critical object.
    Then a critical object $o \in O$ is chosen at random, and $m$ instances of this object are chosen at random.
    The polygon annotations (a polygon border) for this object are obtained from the COCO dataset as shown in Figure~\ref{fig:bbox}.
    Using these annotations, either a removal or color-inversion operation is applied to create the mutant image.

    \subsection{Object Removal and In-painting}
    After the object instance is selected, it is removed from the image by replacing all pixel values by 1 (white).
    This masked image is then input to a GAN-based image inpainting network~\cite{yu2018generative} that fills up this pixels in the mask.
    This makes the image photorealistic.
    This network is one of the best available off-the-shelf blind image inpainting models, and is trained on the ImageNet~\cite{deng2009imagenet}.
    The masked image could also be used as the mutant image however we prefer to use photorealistic images for two main reasons.
    First, masked images do not lie in the same distribution as natural images, and secondly, the mask boundary may give clues to the network about the the shape or outline of the missing object.

    \subsection{Color Inversion Process}
    For mutation that involves a change in the color of the object, we perform a simple pixel-wise color inversion operation on each pixel in the mask to get the mutant image as shown in Figure~\ref{fig:col}.
    This is to ensure that we do not use any prior knowledge about valid colors of a specific object.
    For instance, bananas can typically be yellow, green, or black. 
    However, if we only change the color or a banana to one of these three colors, we would be using domain knowledge and inadvertently introducing answers from the test set, defeating the purpose of OOD generalization.
    Although the simple inversion process can introduce unnatural colors like blue bananas, it forces the model to understand colors in the image to answer the question instead of simply answering from linguistic priors (such as the memorized knowledge that bananas can be green, yellow, or black).

    \subsection{Answer Generation}
    The new answers are generated based on the type of question.
    For \textbf{yes-no questions}, if all instances of the object are removed then the answer changes from yes to no. 
    If only some instances are removed or if the object is non-critical, the answer remains the same.
    For \textbf{number questions}, if $m$ instances of a critical object are removed, the answer changes from $n$ to $n-m$, else the answer remains the same.
    For \textbf{color-based questions} we convert the answer color to their HEX value using Webcolors \footnote{\url{https://pypi.org/project/webcolors/}}, invert the value, and find the color in CSS-21 colors closest to this value to generate the new answer.
       

\begin{table*}[t]
    \centering
    \resizebox{\textwidth}{!}{
    \begin{tabular}{@{}lllll@{}}
        \toprule
        \textbf{Mutation} & $\mathbf{Q}$ & $\mathbf{A}$ & $\mathbf{Q_{mutant}}$ & $\mathbf{A_{mutant}}$ \\
        \toprule
        \multirow{4}{*}{Negation} & Is this bread? & yes & Is this not bread & no \\
         & What is the color of the woman's shirt? & black & What is not the color of the woman's shirt? & white\\
         & Are there deciduous trees? & no & Are there no deciduous trees? & yes \\
         & Is there a boy? & no & Is the no boy? & yes \\
        \midrule
        \multirow{3}{*}{Adversarial} & Who is riding the boat? & man & Who is riding the desk & \textit{``can't say"}\\
         & How big is the plane? & large & How big is the book? & \textit{``size"} \\
         & How many pillows are on the bed? & four & How many pillows are on the table? & \textit{``number"} \\
        \midrule 
        \multirow{4}{*}{Masking} & What type of drink is being displayed? & wine & What type of [MASK] is being displayed? & \textit{``beverage"} \\
         & How many bins? & two & How many [MASK] ? & \textit{``number" }\\
         & What is the green stuff on the sandwich? & lettuce & What is the green stuff on the [MASK]? & \textit{``food"} \\
        \bottomrule
    \end{tabular}
    \caption{Examples of three types of question mutation with new answers}
    \label{tab:negation}
    }
\end{table*}
\begin{table*}[H]
    \centering
    \small
    \begin{tabular}{@{}ll @{}}
        \toprule
        \textbf{Category} & \textbf{Member Answers}\\
        \toprule
        time of day & afternoon, dusk, sunset, sunrise, night, daytime, evening, morning, dawn\\
        weather     & rainy, sunny, snowing, cloudy, windy, storm, blurry, fog, dust-storm, tornado \\
        action      & playing, cutting, talking, dancing, singing, hugging, waving, working, standing, sitting \\
        emotion     & angry, curious, tired, happy, bored, surprise, confused, sad \\
        furniture   & table, chair, desk, couch, sofa, bed, ottoman, barstool, seat, bench \\
        material    & granite, wood, brick, glass, stone, oak, concrete, asphalt, plaster \\
        country     & taiwan, france, canada, indonesia, australia, india, britain, hong kong \\
        person      & man, woman, child, kid, boy, people, baby, lady, girl, adult, male, female, players \\
        \bottomrule
    \end{tabular}
    \caption{Examples of answer categories and member answers per category}
    \label{tab:categories}
\end{table*}
\begin{figure*}[t!]
    \centering
    \includegraphics[width=\linewidth]{ 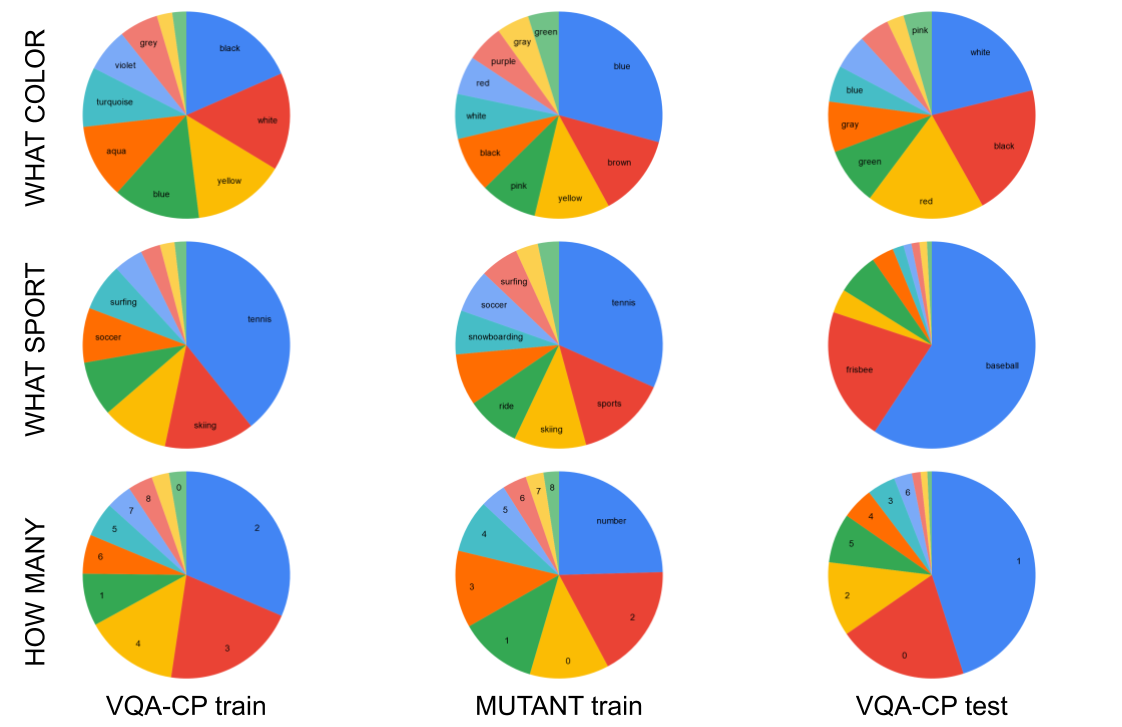}
    \caption{The distribution of answers by question types for VQA train and Mutant compared with VQA-test}
    \label{fig:a_distrib}
\end{figure*}

\section{Question Mutant Generation Process}
For generating question mutants, we use three operators: negation, substitution by antonyms or adversarial words, and masking critical words. 

    \subsection{Negation}
    
    For yes-no questions and color-based questions, we use a template-based negation technique that puts a negative word such as \textit{``not"} or \textit{``no"} before a preposition, noun phrase, or verb.
    For instance ``Is this chair broken?" is negated to ``Is this chair not broken?".
    We show examples of negation in Table~\ref{tab:negation}.
    Negation simply flips the answer from yes to no or no to yes.
    
    \subsection{Adversarial Words and Masking}
    Another form of question mutation is substituting object-words with their adversarial words.
    To do so, we create a list of all object words and their synonyms and use BERT~\cite{devlin2018bert} similarity to rank the most similar words.
    To replace a word, we chooser the most similar word which is not present in the image.
    The third type of mutation is masking, where a critical object word is removed from the question and replaced with the token ``MASK".
    
    For both these types of mutations, determining the correct answer in some cases is not possible as can be seen from examples in Table~\ref{tab:negation}.
    Thus we use the broad category as the answer. 
    For instance, when a question such as \textit{``How big is the book"} is replaced with either \textit{``How big is the plane"} or \textit{``How big is the [MASK]"}, it is clear that the question is about the size of an object.
    Thus we annotate this question with this broad category ``size" as the answer.
    In other cases, where even a broad category cannot be ascertained, the answer is replaced with ``can't say" or "don't know".

    To generate answer clusters and representative answer categories, we extract Glove~\cite{pennington2014glove} word vectors for each answer phrase/word using Spacy. 
    We use k-means clustering~\cite{lloyd1982least} with Euclidean distance metric and with varying number of $K$. 
    We manually tune the number of clusters till we observe a clear set of categories appear at $K=50$.
    We then manually annotate the category names.


\section{Dataset Analysis}
Here we provide dataset analysis in terms of distribution of answers by question-type, number of samples for each type of mutation, and the final distribution of the dataset in terms of answer-type.

\begin{table}[t]
    \centering
    \small
    \begin{tabular}{| c | c | c |}
        \hline
        \textbf{Category} & \textbf{VQA-CP (\%)} & \textbf{Mutant (\%)}\\
        \hline 
        Yes/No & 41.86 & 47.88 \\
        Number & 11.91 & 13.64 \\
        Other  & 46.23 & 38.48 \\
        \hline
    \end{tabular}
    \caption{Distribution of samples in the dataset by answer type}
    \label{tab:ans_type}
\end{table}
    \subsection{Distribution by Question Type}
    We show the distribution of answers per question type in Figure~\ref{fig:a_distrib} for three categories \textit{``How many", ``What sport", and ``What color"} for the top-10 answers.
    It can be seen that the distribution is distinct from the test data and close to the VQA-CP train data apart from the introduction of categorical answers such as \textit{``number" and ``sports"} during question mutation.
    Our mutation method does not leak information about answers from test set to train set.

        \begin{table}[t]
    \centering
    \small
    \begin{tabular}{@{}cc@{}}
        \toprule
        \textbf{Mutation Category} & \textbf{Number of Samples}\\
        \toprule
        Object Removal  & 159,899\\
        Color Change    & 30,759 \\
        Negation        & 237,611 \\
        Adversarial Substitution & 146,814 \\
        Word Masking & 104,666\\
        \bottomrule
    \end{tabular}

    \caption{Distribution of generated mutant samples by category of mutation}
    \label{tab:mutant_type}
\end{table}

    \subsection{Distribution by Mutation Type}
    Table~\ref{tab:mutant_type} shows the number of samples generated by each type of mutation.

    \subsection{Distribution by Answer Type}
    There are three answer types in both VQA-CP and Mutant datasets: \textit{yes/no, number, and other}.
    Creation of mutant samples leads to a small change in the distribution as shown in Table~\ref{tab:ans_type}.

\end{document}